%% file: root.tex
\documentclass[letterpaper, 10 pt, conference]{ieeeconf}  

\IEEEoverridecommandlockouts                              

\usepackage{graphicx}
\usepackage{graphics} 
\usepackage{amsmath} 
\usepackage{amssymb}  
\usepackage{xcolor}
\usepackage{xspace}
\usepackage{subcaption} 
\usepackage{booktabs}
\usepackage{hyperref}

\newcommand{\vspacebeforefigure}[0]{\vspace{2mm}}
\newcommand{\vspacebeforesubcaption}[0]{\vspace{-5mm}}

\newcommand{\vspaceafterfigure}[0]{\vspace{-4mm}}  
\input{00_symbols}

\title{\LARGE \bf
\rmfd{}: A Combined Reachability and Inverse Reachability Map for Common 6-/7-axis Robot Arms by Dimensionality Reduction to 4D
}

\author{
    Martin Rudorfer
    \thanks{
        The author is with the Department of Applied AI \& Robotics and the Aston Centre for Artificial Intelligence Research and Application (ACAIRA), Aston University, Birmingham, UK.
        {\tt\small m.rudorfer@aston.ac.uk}
    }
}

\begin{document}
\maketitle
\thispagestyle{empty}
\pagestyle{empty}

\bstctlcite{IEEEexample:BSTcontrol}

\begin{abstract}

Knowledge of a manipulator's workspace is fundamental for a variety of tasks including robot design, grasp planning and robot base placement.
Consequently, workspace representations are well studied in robotics.
Two important representations are reachability maps and inverse reachability maps.
The former predicts whether a given end-effector pose is reachable from where the robot currently is, and the latter suggests suitable base positions for a desired end-effector pose.
Typically, the reachability map is built by discretizing the 6D space containing the robot's workspace and determining, for each cell, whether it is reachable or not.
The reachability map is subsequently inverted to build the inverse map.
This is a cumbersome process which restricts the applications of such maps.
In this work, we exploit commonalities of existing six and seven axis robot arms to reduce the dimension of the discretization from 6D to 4D.
We propose Reachability Map 4D (\rmfd{}), a map that only requires a single 4D data structure for both forward and inverse queries.
This gives a much more compact map that can be constructed by an order of magnitude faster than existing maps, with no inversion overheads and no loss in accuracy.
Our experiments showcase the usefulness of \rmfd{} for grasp planning with a mobile manipulator.

\end{abstract}

\section{INTRODUCTION}
\input{01_introduction}

\section{RELATED WORKS}
\input{02_related_works}

\section{\rmfd{}}
\input{03_map}

\section{EXPERIMENTS}
\input{04_experiments}

\section{CONCLUSION}
\input{05_conclusion}

\addtolength{\textheight}{-19cm}   






\bibliographystyle{IEEEtran}
\bibliography{bibliography}

\end{document}

%% file: 00_symbols.tex
\newcommand{\rmfd}{RM4D}

\newtheorem{assumption}{Assumption}

\def\p{\Vec{p}}
\def\rx{\Vec{r}_x}
\def\ry{\Vec{r}_y}
\def\rz{\Vec{r}_z}
\def\rzx{r_{z,x}}
\def\rzy{r_{z,y}}

\def\z{\Vec{z}}

\def\xc{x^*}
\def\yc{y^*}

\def\T{\mathcal{T}}
\def\TEE{\T^{TCP}_{base}}
\def\TEEw{\T^{TCP}_{world}}

\def\R{\mathbb{R}}
\def\S{\mathbb{S}}

\DeclareMathOperator{\atantwo}{atan2}


%% file: 01_introduction.tex
When we humans attempt to grasp an object, we know intuitively whether we will be able to reach it from our current position or not.
For a robot arm, this is determined by its workspace.
An easily accessible representation of the workspace is useful for a variety of tasks including grasp planning, robot design, and robot base placement.
However, finding a good representation of this workspace is non-trivial as we can generally only probe points within the workspace using forward and inverse kinematics.
One such representation is a so-called reachability map, also capability map, as first proposed by Zacharias et al. in~\cite{Zacharias2007} and~\cite{Zacharias2013}.
The map is essentially a data structure obtained by discretization of the 6D~space containing the robot's workspace.
During an offline stage, the map is built by determining for each cell whether it is considered reachable or not.
Then, during the online stage, the map can be queried with a pose of the Tool Center Point (TCP), the cell it falls into is identified, and the reachability value can be looked up.

\begin{figure}
\vspacebeforefigure
\centering
    \begin{subfigure}[b]{0.47\linewidth}
    \centering
    \includegraphics[width=\textwidth,trim={3cm 4.5cm 3cm 0},clip]{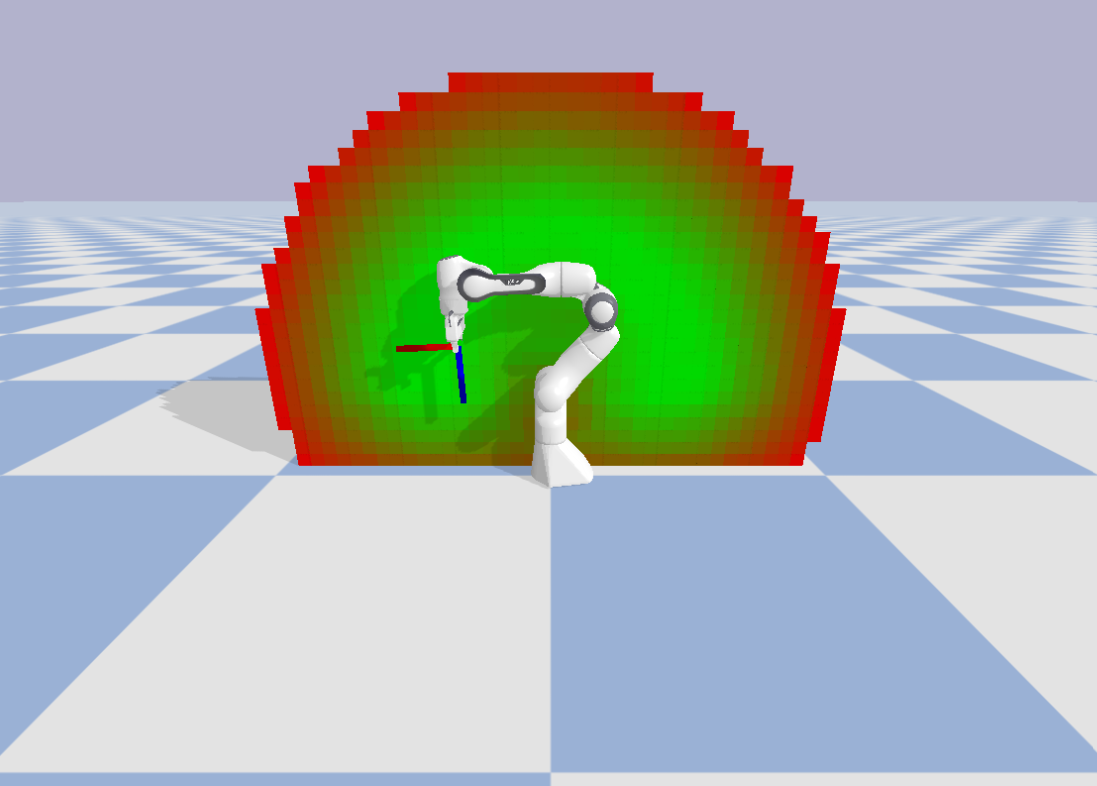}
    \caption{\label{fig:forward_map}}
    \end{subfigure}
\quad
    \begin{subfigure}[b]{0.47\linewidth}
    \centering
    \includegraphics[width=\textwidth,trim={0 0 0 1.17cm},clip]{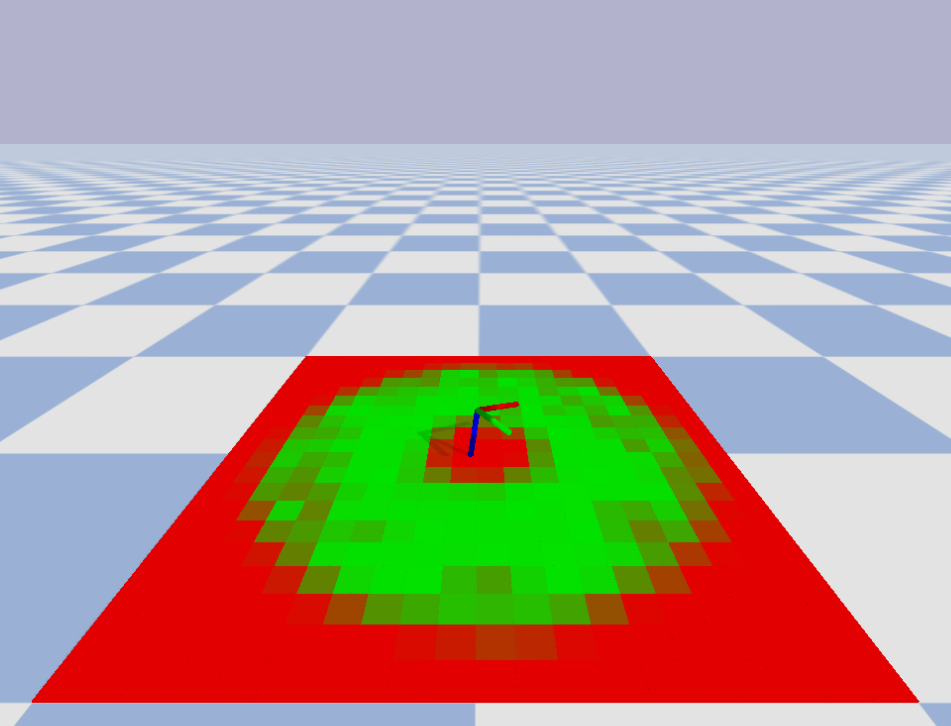}
    \caption{\label{fig:inverse_map}}
    \end{subfigure}
\vspacebeforesubcaption
    \caption{
        \rmfd{} has a single data structure that can be queried as (a) reachability map and (b) inverse reachability map.
        The coordinate system indicates the desired TCP pose, with the blue axis being the approach vector.
        }
    \label{fig:forward-inverse-map}
    \vspaceafterfigure
\end{figure}

Due to the high dimensionality, a trade-off arises between the resolution of the map and both the memory required to store the map as well as the time required to build it.
Although there are several more recent works on capability maps, the same underlying structure as in~\cite{Zacharias2013} or a similar 6D~structure is used by most of them~\cite{Makhal2018, Yao2024}.
There are some works on reducing the time needed for building the map~\cite{Han2021}.
Only few works focus on finding a more compact representation.
One option is to disregard the in-plane rotation of the end-effector, which allows to reduce the number of dimensions from six to five~\cite{Quan2022}.
With \rmfd{}, we propose a dimensionality reduction from six to four dimensions, which significantly reduces both memory required as well as the construction time of the map while preserving the accuracy.

So far, we have discussed the question of whether a certain TCP pose is reachable, given that the robot's base is fixed.
We can also ask the inverse question: given a certain TCP pose, where should the robot's base be so that it can reach it?
This is important for robot base placement, e.g., in the design of workcells for fixed arms or during planning for mobile manipulation tasks~\cite{KlugeWilkes2021}.
A common approach is to first create a capability map and then invert it, such as in~\cite{Vahrenkamp2013} or~\cite{Makhal2018}.
This usually involves creating a second 6D data structure and going through all the voxels of the capability map to invert the poses and fill the inverse capability map.
In \rmfd{}, we store a canonical base position, which allows us to query the same 4D data structure for both reachability and inverse reachability without any inversion overheads (see Figure~\ref{fig:forward-inverse-map}).

In this work, we propose \rmfd{}, a combined reachability and inverse reachability map for common six-axis and seven-axis robot arms that only requires a single 4D data structure.
We achieve this by employing a mapping that reduces the dimensionality from 6D to 4D, exploiting commonalities of such arms.
Specifically, we assume that both the first and the last axis can perform a full rotation -- which holds true for a large proportion of arms and \emph{almost} true for most others.
Our experiments show that \rmfd{}:
\begin{itemize}
    \item can be constructed by an order of magnitude faster than other maps,
    \item reduces memory requirements by an order of magnitude due to its compact representation,
    \item reaches state-of-the-art accuracy even when the assumptions are not fully satisfied,
    \item can be directly queried as both forward and inverse map.
\end{itemize}

We showcase the use of this map for a grasp planning scenario with a mobile manipulator, which involves using the inverse map to find a suitable base position given a large set of grasp candidates, and then using the forward map to subsequently filter reachable grasps.

The code and scripts to reproduce all experiments are publicly available on \href{https://mrudorfer.github.io/rm4d/}{mrudorfer.github.io/rm4d/}.

%% file: 02_related_works.tex
\label{sec:related-works}

Estimating a robot's workspace has a long-standing research history due to its importance for robot design and work cell design.
Earlier works mostly estimated the workspace as boundary or simple shapes in 2D and 3D~\cite{Castelli2008,Cao2011}.
However, this simplification demands to define several types of workspaces.
E.g., the reachable workspace, which contains all points that are reachable in at least one orientation, or the dexterous workspace, which contains only those points that are reachable in all orientations~\cite{Castelli2008}.

Zacharias et al.~\cite{Zacharias2013} demonstrated how a capabilty map can model the types of workspaces more holistically.
They discretize the 6D space by defining a 3D voxelgrid that encapsulates the workspace.
The orientation is separated into the approach direction and the in-plane rotation.
They inscribe a sphere into each voxel and evenly distribute a number of points on the surface of the sphere.
The vectors from each surface point to the center of the sphere make up the approach directions.
For each resulting approach direction, they then use a number of in-plane rotations to finalize the discretization of $SO(3)$.
The resulting data structure is a 5D array, where the two DOF of the approach direction, encoded by the sphere point index, are stored in one dimension of the array.

This is a typical discretization strategy for capability maps and has been utilized by a number of other works in this exact or a similar manner~\cite{Yao2024}.
In~\cite{Porges2015}, they also inscribe spheres into each voxel, but determine the approach directions differently.
In~\cite{Zacharias2007} and \cite{Quan2022}, they exploit the fact that the distribution of reachable points on each sphere would follow certain patterns, and they approximate those using shape primitives to achieve data reduction and simplify querying the map.
Disregarding the in-plane rotation also allows to reduce the dimensionality~\cite{Quan2022}.
Rouleaux~\cite{Makhal2018} uses the same structure as~\cite{Zacharias2013}, but employ an octree to allow for adaptable voxel sizes.
A compact representation is crucial, as it also allows to reduce the effort required to build the map.
Various ways of constructing capability maps based on random sampling and forward kinematics or inverse kinematics have been compared in~\cite{Porges2015}, but finding more efficient strategies is still subject to research~\cite{Han2021}.

For applications like base position planning, an inverse capability map is required.
This is typically built by considering every pose in the capability map as transformation matrix, inverting it, and storing it in an inverse map~\cite{Vahrenkamp2013,Makhal2018,Burget2015}.
In~\cite{Zhang2020}, they store a filtered list of end-effector poses and their inverses explicitly (without any discretization), and they employ a linear search to find suitable base positions to plan a trajectory for a mobile manipulator.
Recently, approaches emerged to model capability maps and their inverses as neural networks~\cite{Kim2021a}.

\rmfd{} has a compact representation in 4D and can be directly used as both forward and inverse map for common six and seven axis robot arms.

%% file: 03_map.tex
\label{sec:rmfd}

In this section we present our Reachability Map 4D.
We explain the core idea for the dimensionality reduction that allows us to store the map in a 4D data structure.
Then we illustrate how the map can be efficiently queried both as forward and inverse reachability map.

\subsection{Dimensionality Reduction}

Without loss of generality, we assume that the robot's base is at the origin and the pose of the TCP with respect to the base is described as follows:
\begin{equation}
    \TEE = 
    \begin{bmatrix}
        | & | & | & | \\
        \rx & \ry & \rz & \p \\
        | & | & | & | \\
        0 & 0 & 0 & 1
    \end{bmatrix}
    \in SE(3),
\end{equation}
where $\p$ is the TCP~position and $\rz$ is the approach vector (blue in our figures).
To achieve a dimensionality reduction from 6D to 4D, we make the following two assumptions:
\begin{itemize}
\item \begin{assumption}
    \label{assumption-wrist}
    The last wrist joint can rotate around 360~degrees, and its axis of rotation is in line with the approach vector~$\rz$.
\end{assumption} 
\item \begin{assumption}
    \label{assumption-base}
    The base can rotate around 360~degrees.
\end{assumption}
\end{itemize}

\begin{table}
\vspacebeforefigure
    \centering
    \caption{Limits for the first/last joints of exemplary 6/7-axis robot arms.}
    \begin{tabular}{c|c|c}
        Robot Arm & Range Base & Range Wrist\\
        \hline
        Kinova Gen3~\cite{Kinova} & infinite & infinite \\
        Universal Robots~\cite{UniversalRobot} & $\pm 360^\circ$ & $\pm 360^\circ$/infinite \\ 
        Kuka LBR iiwa~\cite{Kuka} & $\pm 170^\circ$ & $\pm 175^\circ$\\
        Franka Panda~\cite{Franka} & $\pm 166^\circ$ & $\pm 166^\circ$ \\
        Franka Research 3~\cite{Franka} & $\pm 157^\circ$ & $\pm 172^\circ$\\
        Kinova Gen3lite~\cite{Kinova} & $\pm 155^\circ$-$160^\circ$ & $\pm 155^\circ$-$160^\circ$ \\
    \end{tabular}
    \label{tab:robot_joint_limits}
    \vspaceafterfigure
\end{table}

Table~\ref{tab:robot_joint_limits} shows the joint limits for some common robot arms, and we notice that these assumptions are only fully satisfied for some of the robots.
But let us proceed for now, and later on in our experiments we will investigate how the accuracy of the map behaves when those assumptions are not fully satisfied.

\begin{figure}
\vspacebeforefigure
\centering
    \begin{subfigure}[b]{0.42\linewidth}
    \centering
    \includegraphics[height=3.5cm, trim={0cm 0cm 3cm 5cm}, clip]{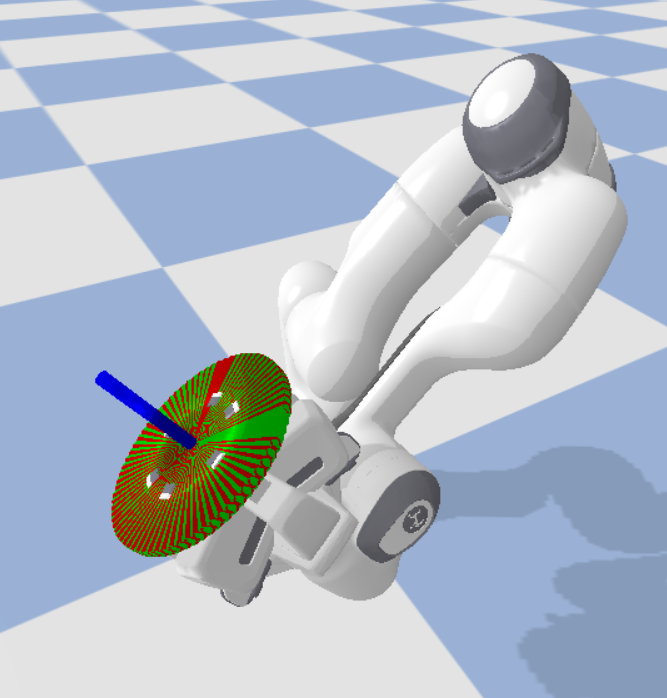}
    \caption{\label{fig:rotation-approach-vec}}
    \end{subfigure}
\quad
    \begin{subfigure}[b]{0.52\linewidth}
    \centering
    \includegraphics[height=3.5cm]{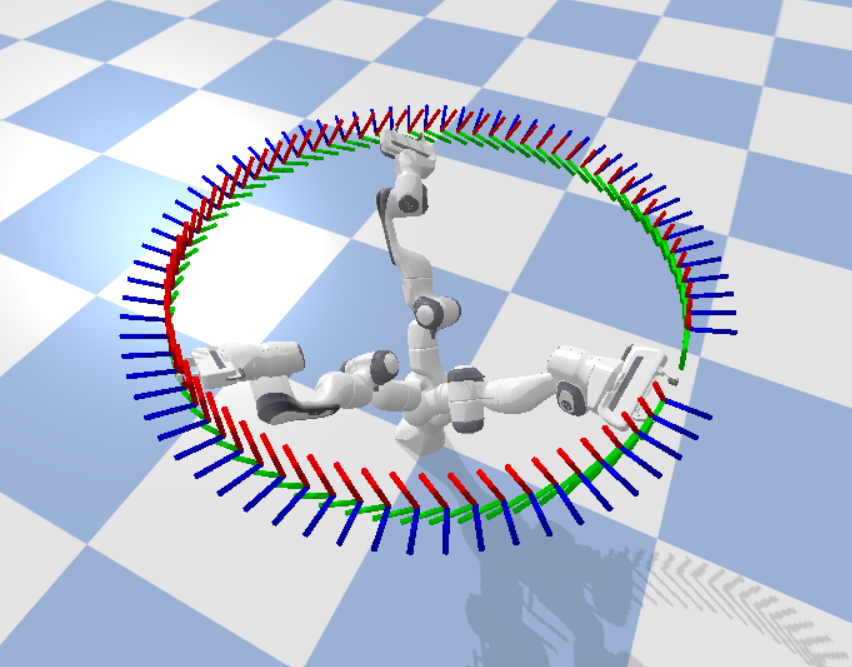}
    \caption{\label{fig:rotation-base}}
    \end{subfigure}
\vspacebeforesubcaption
    \caption{
        Illustration of our two assumptions: 
        (a) Rotating the last wrist joint only affects the in-plane rotation.
        The TCP position and the approach vector (blue) remain constant.
        (b) Rotating the base gives an arc/circle.}
    \vspaceafterfigure
\end{figure}

Looking at Figure~\ref{fig:rotation-approach-vec}, we see that a rotation of the last wrist joint does not change the TCP position~$\p$ nor the approach vector~$\rz$.
Under Assumption~\ref{assumption-wrist}, the position of the last wrist joint has no effect on the reachability.
It is hence sufficient to describe the TCP pose as $(\p, \rz)$.

Next, in Figure~\ref{fig:rotation-base}, we can see that rotating the base gives a set of TCP poses whose positions lie on a continuous arc, which under Assumption~\ref{assumption-base} is a full circle.
As determined above, the illustrated in-plane rotation is arbitrary and can be disregarded.
We would now like to map the set of all these poses to a single element in 4D.

\begin{figure*}
\vspacebeforefigure
\centering
    \begin{subfigure}[b]{0.3\textwidth}
    \centering
    \includegraphics[width=\textwidth]{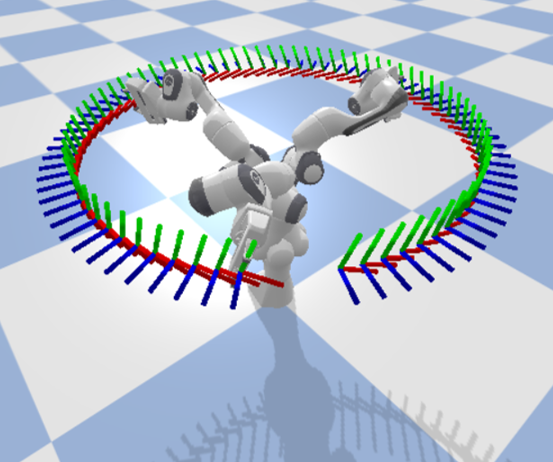}
    \caption{\label{fig:tf_step0}}
    \end{subfigure}
\quad
    \begin{subfigure}[b]{0.3\textwidth}
    \centering
    \includegraphics[width=\textwidth]{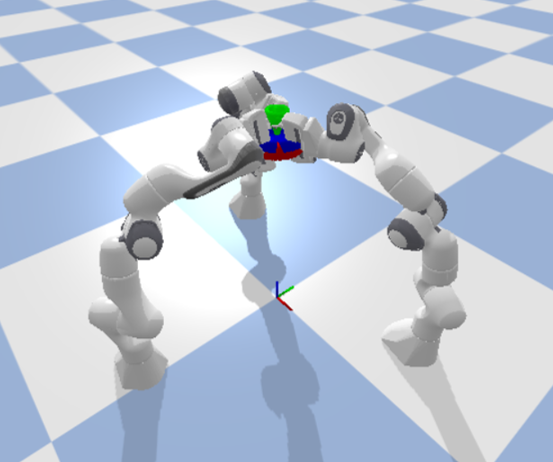}
    \caption{\label{fig:tf_step1}}
    \end{subfigure}
\quad
    \begin{subfigure}[b]{0.3\textwidth}
    \centering
    \includegraphics[width=\textwidth]{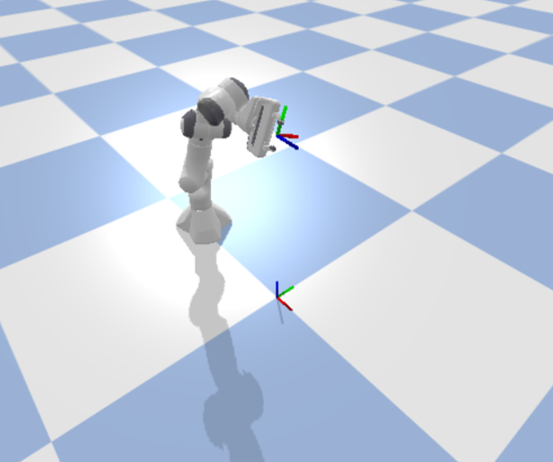}
    \caption{\label{fig:tf_step2}}
    \end{subfigure}
\caption{
    The steps in our canonical transformation, allowing the dimensionality reduction:
    (\subref{fig:tf_step0}) TCP~frames for different angles of the base joint,
    (\subref{fig:tf_step1}) TCP~frames are centered at origin,
    (\subref{fig:tf_step2}) TCP~frames are rotated such that the approach vector~$\rz$ (blue) is contained in the $x^{(+)}z$ half-plane.
    As a result, we get a canonical base position $(\xc, \yc)$ that is identical for all configurations in (\subref{fig:tf_step0}).
    With this, we can map the 6D pose to the 4D vector $(p_z, \theta, \xc, \yc)$, where $p_z$ is the z-coordinate of the TCP and $\theta$ the angle between $\rz$ and the world z-axis (up).
}
\vspaceafterfigure
\label{fig:tf_steps}
\end{figure*}

We define the mapping $f: SE(3) \rightarrow \R^3 \times \S^1$ as:
\begin{equation}
    f(\TEE) = (p_z, \theta, \xc, \yc),
    \label{eq:mapping}
\end{equation}
where $p_z$ is the z-coordinate of~$\p$, $\theta = \sphericalangle (\rz, \z)$ is the angle between the approach vector~$\rz$ and the world's z-axis (up), and $(\xc, \yc)$ is a canonical base position.
Both $p_z$ and $\theta$ are identical for all the poses in the set.
The steps to get $(\xc, \yc)$ are illustrated in Figure~\ref{fig:tf_steps}.
First, we translate $\p$ onto the $\z$ axis, which gives an intermediate base position at $[-p_x, -p_y, 0]^T$  (Figure~\ref{fig:tf_step1}).
Second, we rotate $\rz$ around $\z$ such that it lies in the $x^{(+)}z$ half-plane (Figure~\ref{fig:tf_step2}).
The rotation angle can be determined based on the 2D-projection.
Specifically, we are looking for the angle~$\psi$ between the x-axis $[1, 0]^T$ and $[\rzx, \rzy]^T$, the first two components of~$\rz$.
The 2-argument-arctangent provides $\psi \in [-\pi, \pi]$ as follows:
\begin{equation}
    \psi = \atantwo(\rzy, \rzx)
    \label{eq:psi}
\end{equation}
We then apply the inverse rotation, which aligns the approach vector with the $x^{(+)}z$ half-plane, to the intermediate base position and yield the canonical base position:
\begin{equation}
    \begin{bmatrix}
        \xc \\ \yc \\
    \end{bmatrix}
    =
    \begin{bmatrix}
        \cos{\psi} & \sin{\psi} \\
        -\sin{\psi} & \cos{\psi} \\
    \end{bmatrix}
    \begin{bmatrix}
        -p_x \\ -p_y \\
    \end{bmatrix}
    \label{eq:rotate}
\end{equation}
This completes the values $(p_z, \theta, \xc, \yc)$, which form the four dimensions of our 4D capability map.

\subsection{Data Structure of the Map}

\rmfd{} stores the four values resulting from the mapping in Equation~\ref{eq:mapping} in a discretized form as a 4D array.
The range of the canonical base position $(\xc, \yc)$ is identical to the range of the TCP positions $(p_x, p_y)$ and can be determined based on the maximum horizontal reach $r_{xy}$ of the robot, such that $\xc$ and $\yc$ lie in the interval $[-r_{xy}, +r_{xy}]$.
Similarly, the range for $p_z$ is determined by the maximum vertical reach $r_z$.
We assume the robot to be on a ground plane and hence only record $p_z \in [0, r_z]$, but the range can be expanded to negative values if required by the application.
This results in a voxelgrid that encapsulates the workspace of the robot.
We choose a uniform edge length of $l_c$ for all voxels, and the sets of indices $\mathcal{N}_{xy}$ and $\mathcal{N}_{p_z}$ are determined accordingly to cover the range.
The angle $\theta$ is in the range of $[0, \pi]$ and discretized using a step of $\Delta_\theta$.
Let $D_P$ be a discretization function that uses the above mentioned parameters $P=(r_{xy}, r_z, l_c, \Delta_\theta)$ to identify the map indices:
\begin{equation}
    i_{p_z}, i_\theta, i_{\xc}, i_{\yc} = D_P(p_z, \theta, \xc, \yc)
    \label{eq:indices}
\end{equation}
Each element of the underlying 4D array $\mathcal{M}$ stores a binary reachability value~$r \in \{0, 1\}$ that can be retrieved using these indices.
\begin{equation}
    r = \mathcal{M}[i_{p_z}, i_\theta, i_{\xc}, i_{\yc}]
    \label{eq:reach}
\end{equation}

\subsection{Querying the Map}

The map can be queried both as forward and inverse map.
For the forward query, we are given a TCP pose~$\TEE$ and would like to determine its reachability.
We simply apply the mapping from Equation~\ref{eq:mapping} to identify the 4D values and then use Equations~\ref{eq:indices} and \ref{eq:reach} to determine the corresponding indices and look up the reachability value in the array.

For the inverse query, we would like to return all suitable base poses from which a given $\TEEw$ can be reached.
We utilize the mapping in Equation~\ref{eq:mapping} to determine only $p_z$ and $\theta$, and Equation~\ref{eq:indices} gives us their corresponding indices $i_{p_z}, i_\theta$.
This can be done because the calculations are independent for each variable.
We can then index the underlying array to obtain the corresponding 2D slice of reachability values for the different base positions.
We only keep indices of cells with a reachability value of~1.
\begin{equation}
    B^i = \{i_{\xc}, i_{\yc} \in \mathcal{N}_{xy} | \mathcal{M}[i_{p_z}, i_\theta, i_{\xc}, i_{\yc}] = 1 \}
\end{equation}

With the inverse of our discretization function, we can obtain the corresponding canonical base positions at the center of each indexed cell:
\begin{equation}
    B^* = \{\xc, \yc = D_P^{-1}(i_{\xc}, i_{\yc}),~\forall i_{\xc}, i_{\yc} \in B^i \}
\end{equation}

We now employ the inverse mapping $f^{-1}$ to obtain the actual base positions $(x_B, y_B)$ from the canonical base positions $(\xc, \yc)$.
This is done by calculating the angle $\psi$ as in Equation~\ref{eq:psi} and rotating back to reverse Equation~\ref{eq:rotate}.
Finally, we need to add the TCP position $(p_x, p_y)$.
The steps are formalized in the following equation:
\begin{equation}
        \begin{bmatrix}
        x_B \\ y_B \\
    \end{bmatrix}
    =
    \begin{bmatrix}
        \cos{\psi} & -\sin{\psi} \\
        \sin{\psi} & \cos{\psi} \\
    \end{bmatrix}
    \begin{bmatrix}
        \xc \\ \yc \\
    \end{bmatrix}
    +
    \begin{bmatrix}
    p_x \\ p_y \\
    \end{bmatrix}
\end{equation}

This is done for all $(\xc, \yc) \in B^*$ and gives us the desired set of base positions $B$, from which the TCP pose $\TEEw$ can be reached.

%% file: 04_experiments.tex
With our experiments we aim to show the characteristics of \rmfd{} with regards to memory requirement, construction time of the map, and accuracy of the map, particularly when the two core assumptions are violated.
We further show how the map can be used as both forward and inverse map for a grasp planning scenario.

\subsection{Experiment Setup}

Our implementation is based on Python, using pyBullet~\cite{Coumans2021} as library for forward and inverse kinematics as well as collision detection.
We evaluate for a 6-DOF UR5e and a 7-DOF Franka Panda robot.
As illustrated in Table~\ref{tab:robot_joint_limits}, the UR5e can do two full rotations with both its first and last axis and hence fully satisfies our assumptions.
The Franka Panda robot has a smaller joint range and is investigated as an exemplary arm that violates both assumptions.

We compare our \rmfd{} to two baseline capability maps.
Firstly, the one by Zacharias et al.~\cite{Zacharias2013}, as described in Section~\ref{sec:related-works}, with 200~sphere points per voxel and 12~in-plane rotations.
Secondly, we use their same structure, but disregard the in-plane rotation, which makes use of our first assumption to reduce the map from 6D to 5D.
We call this variant \emph{Zacharias (5D)}.
The voxel size is $l_c=5cm$ for all maps, and the rotation step for \rmfd{} is $\Delta_\theta=5^\circ$.
The resulting sizes of the respective underlying data structures are displayed in Table~\ref{tab:sizes}.
Already, we can see that \rmfd{} is a much more compact representation of the workspace.

\begin{table}
    \centering
    \caption{Comparison of the map sizes for the Franka Panda robot.}
    \label{tab:sizes}
    \begin{tabular}{c|c}
        Map Type & No of Cells \\
        \hline
        Zacharias et al. \cite{Zacharias2013} & 114,307,200 \\
        \hline
        Zacharias et al. 5D & 9,525,600 \\
        \hline
        \rmfd{} (ours) &  1,714,608 \\
    \end{tabular}
    \vspaceafterfigure
\end{table}

\subsection{Construction of the Map}

There are various ways to construct capability maps.
In~\cite{Zacharias2013}, they specifically tailor the construction method to their discretization of the orientation, and it combines random sampling with inverse kinematics (IK).
Although we could design a similar strategy for \rmfd{}, we prefer to use identical sampling strategies for both maps to have a fair comparison.
Therefore, we randomly sample configurations from within the robot's joint limits.
If they are collision-free, we use forward kinematics to calculate the TCP pose $\TEE$, determine the corresponding cell in the reachability map, and mark it as reachable.

\begin{figure}
    \centering
    \includegraphics[width=0.48\linewidth,trim={0 0 1.2cm 0},clip]{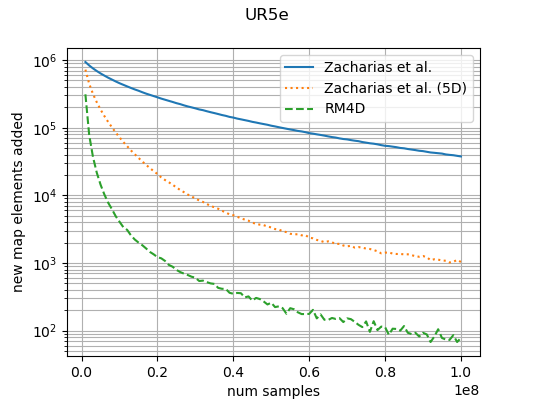}
    ~
    \includegraphics[width=0.48\linewidth,trim={0 0 1.2cm 0},clip]{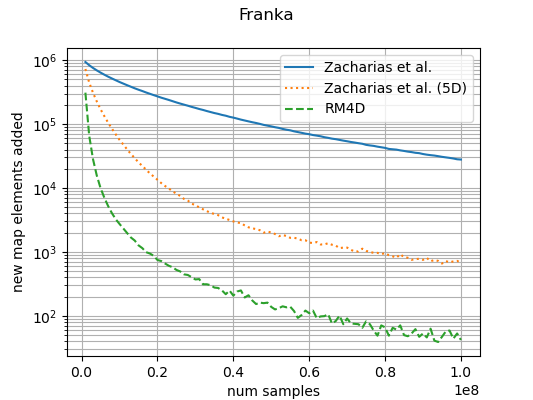}
    \caption{Number of novel map elements visited per 1M samples throughout the map construction.
    The map is sufficiently filled as this value approaches zero.
    \rmfd{} can be constructed with an order of magnitude fewer samples.}
    \label{fig:hits}
    \vspaceafterfigure
\end{figure}

In the limit of the number of samples going towards infinity, we can be sure that all reachable cells have been visited.
Practically, we sample 100M collision-free configurations to build the maps.
There will be cases in which a newly sampled configuration maps to a cell that has already been marked as reachable by a previously sampled configuration.
The longer the construction process continues, the more often this will happen.
We are interested in how the number of new cells visited per 1M samples changes throughout the construction process.
A low number indicates that the map has been mostly filled.
Figure~\ref{fig:hits} shows the plots for the UR5e and the Franka robot. 
They both look very similar.
For \rmfd{}, this value drops to below 10k before reaching 10M~samples, which means that at that stage, 99.9\% of newly sampled configurations map to cells that have already been visited.
This indicates that the map has been fully constructed.
In comparison, Zacharias (5D) drops below 10k after about 25M~samples, and with the added in-plane rotations, it does not reach that point even after 100M samples, indicating that it could be necessary to increase the number of samples even further.
The construction time is directly related to the map sizes, and this experiment confirms the picture painted by Table~\ref{tab:sizes}.

\subsection{Accuracy}

\begin{figure*}
\vspacebeforefigure
\centering
\includegraphics[width=\textwidth]{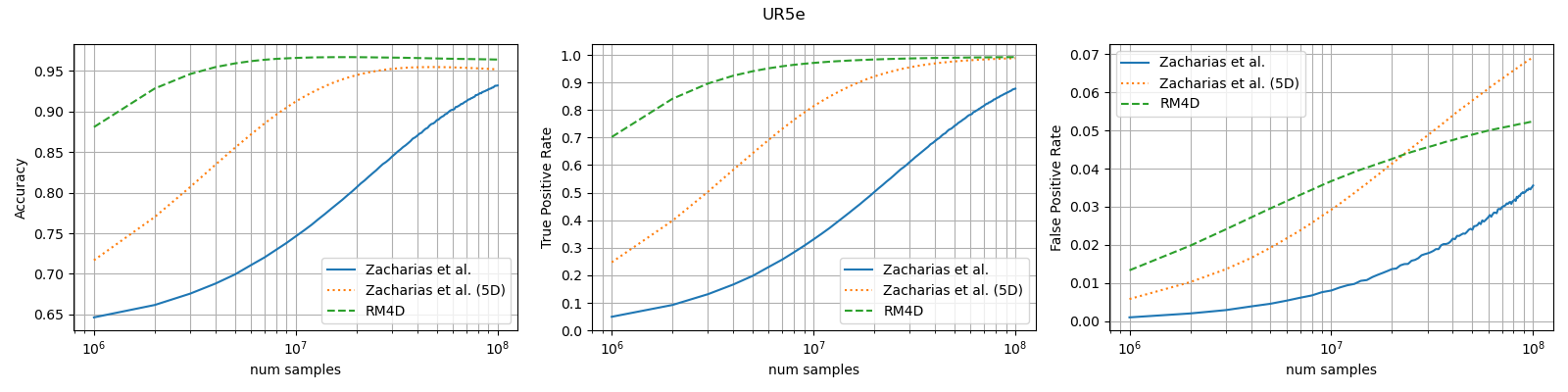}
\quad
\includegraphics[width=\textwidth]{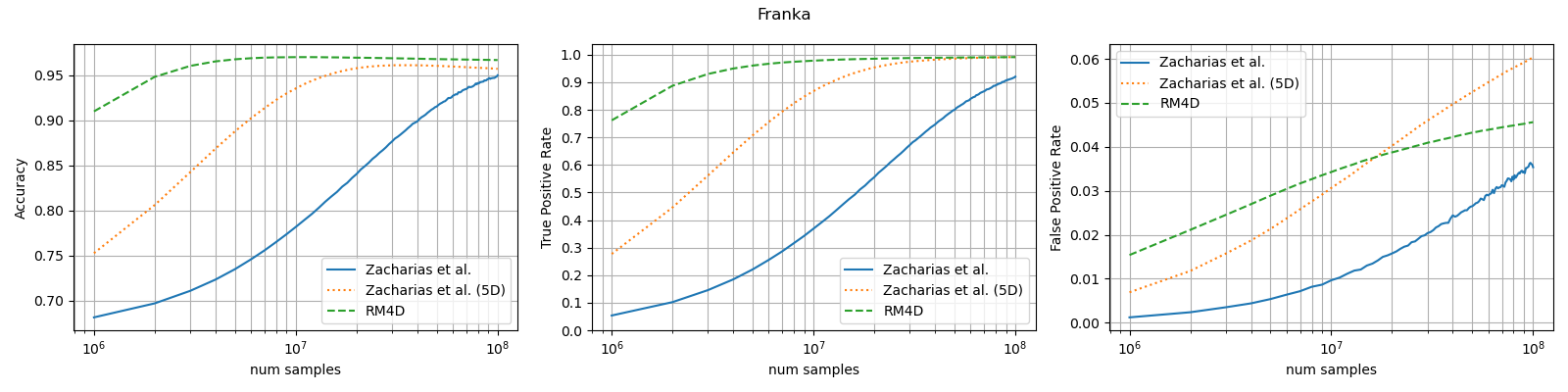}
\caption{
    Accuracy, True Positive Rate (TPR), and False Positive Rate (FPR) throughout the construction of the maps for the UR5e (top) and Franka (bottom).
    \rmfd{} reaches a higher accuracy with fewer samples.
}
\label{fig:accuracy}
\vspaceafterfigure
\end{figure*}

The map is used to predict the reachability of end-effector poses.
We use 1M evaluation poses to test how accurate these predictions are.
The poses are uniformly sampled from within a cylinder that encapsulates the workspace of the robot.
The ground area is defined by the reachable radius of the robot $r_{xy}$, and the height of the cylinder is $r_z$.
The orientation is sampled uniformly using SciPy~\cite{SciPy}.

For each of the evaluation poses, we require the ground-truth reachability information.
We use inverse kinematics (IK) to determine this.
As the success of IK depends on the starting configuration, we perform up to 100~attempts with random initial conditions.
If a collision-free configuration is found that is close enough to the sampled pose, it is considered reachable.
As distance measure we use a combined measure of translational and rotational distance, where we weigh them such that $1mm$ equals $1^\circ$, and we set this threshold to 25.
Roughly a third of the evaluation poses is found reachable.
To get the full picture, we report not only accuracy, but also true positive rate (TPR) and false positive rate (FPR), over the whole construction process.
Note that we now use a logarithmic scale for the number of samples.

Figure~\ref{fig:accuracy} shows the results.
Again, the plots look qualitatively similar for UR5e and Franka.
For \rmfd{}, the accuracy rises above 95\% after 4M and 3M samples, respectively, whereas Zacharias 5D reaches this value at 25M and 15M samples.
The 6D Zacharias map would require more samples to saturate the accuracy.
However, interestingly, even after saturation, the accuracy of Zacharias 5D remains slightly below the accuracy of \rmfd{}.
Looking at the TPR (recall), we notice that both converge to 99\% eventually.
However, the FPR curves look different.
They tell us how many of the non-reachable poses have mistakenly been predicted as reachable.
Surprisingly, Zacharias 5D has a worse FPR than \rmfd{} -- even for the Franka Panda robot which does not satisfy both our assumptions.
We believe that this may be due to the different way of discretizing the orientation.
While the Zacharias maps use 200~bins for two rotational DOF, \rmfd{} uses 36~bins for one rotational DOF, which may resemble a finer resolution, helping to avoid false positives.
This demonstrates that our 4D representation has not only advantages in terms of compactness, but also accuracy.

\begin{figure*}
\vspacebeforefigure
    \centering
    \includegraphics[width=\textwidth]{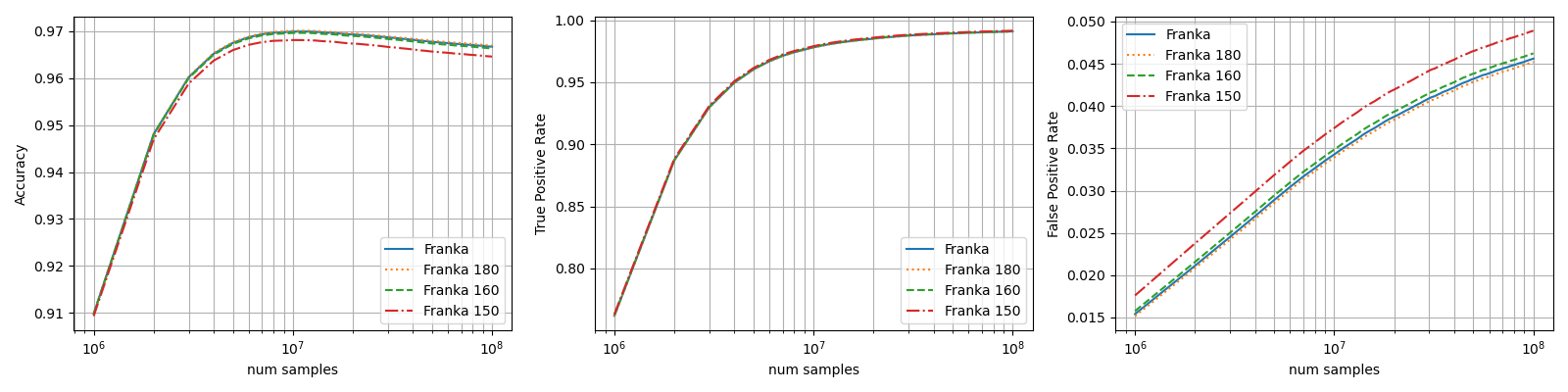}
    \caption{We artificially adjust the range of joints 1 and 7 of the Franka Panda robot (normally $\pm166^\circ$) to gradually violate the assumptions for the dimensionality reduction.
    The accuracy remains high, and only starts dropping slightly when reducing the joint range from $\pm160^\circ$ to $\pm150^\circ$.}
    \label{fig:ablation}
    \vspaceafterfigure
\end{figure*}

It is somewhat surprising that the results are almost the same for the UR5e as well as the Franka Panda, despite the Franka not fully satisfying the assumptions for our dimensionality reduction (see Section~\ref{sec:rmfd}).
If anything, the accuracy for the Franka seems slightly higher than for the UR5e.
Let us investigate further how a violation of the assumptions affects the accuracy.
For a systematic evaluation, we use the same robot and artificially change the limits of the first and last joints in the URDF.
We compare the actual Franka Panda robot, which has a range of $\pm166^\circ$, to versions with $\pm180^\circ$, $\pm160^\circ$, and $\pm150^\circ$.
The results are shown in Figure~\ref{fig:ablation}.
As expected, the recall (TPR) stays the same and only the FPR is affected.
Notably, the FPRs for the versions from $\pm180^\circ$ down to $\pm160^\circ$ are almost identical.
Restricting the Franka to $\pm150^\circ$ shows some more false positives, but not unreasonably many.
The overall accuracy only drops by 0.2\% compared to the $\pm180^\circ$ version.
This demonstrates that the dimensionality reduction performed in \rmfd{} does not cause the accuracy to catastrophically deteriorate when the assumptions are slightly violated.

\subsection{Planning Base Position for Grasp Planning}

\begin{figure}
    \centering
    \includegraphics[width=0.8\linewidth]{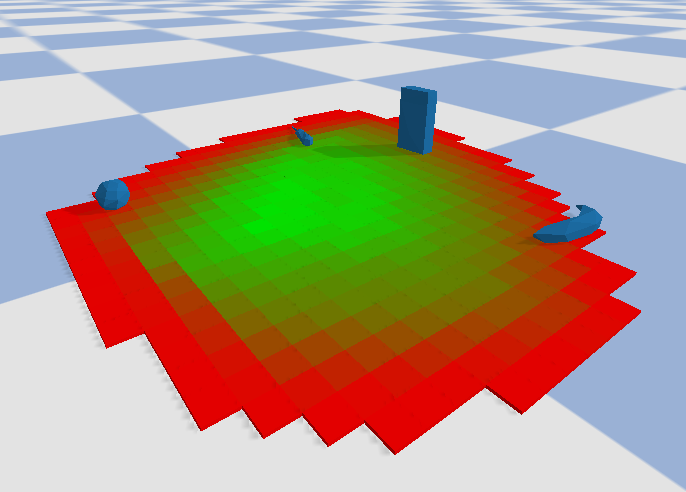}
    \caption{Using \rmfd{} to determine possible base positions from which all four of these objects can be grasped.
    The color encodes the number of reachable grasp candidates.}
    \label{fig:base_positions}
\vspaceafterfigure
\end{figure}

To showcase \rmfd{}'s use for grasp planning, we created a scenario with four objects from the YCB dataset~\cite{Calli2015}.
For each of them, we generated 200~grasp candidates using the BURG Toolkit~\cite{BURG_toolkit_2022}, in total 800~target poses.
We now use \rmfd{} as inverse map, to determine the set of base positions $B$ for each grasp candidate $\TEEw$, and aggregate them in a regular grid for each object separately.
The grid cell with the highest value represents the base position from which most of the grasp candidates for that object are reachable.
We then use the minimum value across all four grids, to ensure that all objects are graspable from the suggested positions.
The result is visualized in Figure~\ref{fig:base_positions}.
Upon selecting the base position with the highest value from the combined grid, we use \rmfd{} as forward map to determine which grasps are reachable from the chosen position.
With our Python implementation, it takes 2.47s to produce the combined grid and 0.02s for the forward queries.
This demonstrates how \rmfd{} can be used to quickly process large amounts of poses for both forward and inverse queries from a single data structure.

%% file: 05_conclusion.tex
In this paper, we presented Reachability Map 4D (\rmfd{}).
Assuming that both the first and last joint of the robot arm can make a full rotation allowed us to employ a dimensionality reduction from 6D to 4D.
We store the information in such a way that the same map can be used for both forward and inverse queries.
This representation of the workspace is much more compact than existing reachability maps, and can hence be built by an order of magnitude faster than other approaches.
There are no overheads in creating an inverse map, and there are no sacrifices in accuracy compared to other maps.
Our experiments have shown that the accuracy remains high even when the assumptions are not fully satisfied.
The map can be used for grasp planning, base position planning, and a variety of other tasks.
Future research could explore how to encode information about the configuration of the robot into \rmfd{}, e.g. manipulability scores or continuous nullspace sections as in~\cite{Yao2024}.
Inspired by~\cite{Sandakalum2022a}, \rmfd{} could also be used for base position planning in the presence of obstacles.